\documentclass{article}

\PassOptionsToPackage{numbers, compress}{natbib}
\usepackage[final]{neurips_2022}

\usepackage[utf8]{inputenc} % allow utf-8 input
\usepackage[T1]{fontenc}    % use 8-bit T1 fonts
\usepackage{hyperref}       % hyperlinks
\usepackage{url}            % simple URL typesetting
\usepackage{booktabs}       % professional-quality tables
\usepackage{amsfonts}       % blackboard math symbols
\usepackage{nicefrac}       % compact symbols for 1/2, etc.
\usepackage{microtype}      % microtypography
\usepackage{xcolor}         % colors
\usepackage{graphicx}

\usepackage{algorithm}
\usepackage{algpseudocode}
\usepackage{bbm}
\usepackage{amsmath}
\usepackage{newtxmath}
\usepackage{bbold}
\usepackage{pifont}
\usepackage{mathtools}
\usepackage{adjustbox}
\usepackage{mathtools}
\usepackage{enumitem}
\usepackage{wrapfig}
\usepackage[table]{colortbl}
\usepackage{pythonhighlight}
\usepackage{listings}

\title{TokenMixup: Efficient Attention-guided \\ Token-level Data Augmentation for Transformers}

\author{%
  Hyeong Kyu Choi\thanks{The first two authors contributed equally}
  \hspace{0.6mm},
  \hspace{1mm}
  Joonmyung Choi$^*$,
  \hspace{0.5mm} Hyunwoo J. Kim\thanks{corresponding author} \\
  Department of Computer Science and Engineering, Korea University\\
  \texttt{\{imhgchoi, pizard, hyunwoojkim\}@korea.ac.kr} \\
}

\begin{document}

\maketitle

% \begin{abstract}
%     Mixup is a commonly adopted data augmentation method for image classification.
%     Recent advances in mixup methods primarily focus on mixing based on saliency.
%     However, many saliency detectors require intense computation and are often burdensome for parameter-heavy transformer models.
%     To this end, we propose TokenMixup, an efficient saliency-based mixup method for transformers.
%     Two variants of Horizontal TokenMixup (HTM) and Vertical TokenMixup (HTM) are introduced, both of which harness self-attention as the saliency detector for efficiency.
%     Interesting properties with respect to curriculum learning and DenseNet-style transformer networks are discussed as well.
%     % Horizontal TokenMixup incorporates a curriculum learning perspective into Mixup, which is a novel attempt in literature.
%     Our experiments show that both methods significantly improve the performance of our baseline models on CIFAR and ImageNet-1K datasets while being efficient than previous methods.
%     Code will be publicized.
% \end{abstract}

\begin{abstract}
    Mixup is a commonly adopted data augmentation technique for image classification.
    Recent advances in mixup methods primarily focus on mixing based on saliency.
    However, many saliency detectors require intense computation and are especially burdensome for parameter-heavy transformer models.
    To this end, we propose TokenMixup, an efficient attention-guided token-level data augmentation method that aims to maximize the saliency of a mixed set of tokens. %???
    % TokenMixup accelerates saliency detection by $\times2$ compared to gradient-based methods, and provides an interesting discussion point regarding curriculum learning with mixup.
    % TokenMixup efficiently computes saliency by $\times2$ compared to gradient-based methods, and reveals an interesting aspect in terms of curriculum learning.
    TokenMixup provides $\times15$ faster saliency-aware data augmentation compared to gradient-based methods.
    Moreover, we introduce a variant of TokenMixup which mixes tokens within a single instance, thereby enabling multi-scale feature augmentation.
    Experiments show that our methods significantly improve the baseline models' performance on CIFAR and ImageNet-1K, while being more efficient than previous methods.
    We also reach state-of-the-art performance on CIFAR-100 among from-scratch transformer models.
    Code is available at \href{https://github.com/mlvlab/TokenMixup}{\texttt{https://github.com/mlvlab/TokenMixup}.}
\end{abstract}
\section{Introduction}
\label{section:introduction}

Various data augmentation methods have been proposed for computer vision tasks.
One of the most successful augmentation methods is mixup~\cite{zhang2017mixup}, which is commonly applied to image classification.
It attempts to augment the input by taking a convex combination of two random instances, and by reassigning the ground truth label correspondingly.
Building on mixup, many subsequent works~\cite{yun2019cutmix, kim2020puzzle, kim2021co, uddin2020saliencymix,verma2019manifold, venkataramanan2021alignmix, hong2021stylemix, dabouei2021supermix} have been introduced.
They focus on mixing instances in a more meaningful way (\textit{e.g.}, saliency-aware mixup, manifold level mixup, submodular diversity maximization \textit{etc.}), while a great majority attempts to perform mixup based on saliency.\\
These saliency-aware mixup methods often entail gradient computation.
To detect salient regions, the input is forward-propagated once and back-propagated to extract the gradient map.
However, such a mechanism tends to be computationally heavy. 
This is especially pertinent to transformer-based models~\cite{vaswani2017attention}, as they generally retain high parameter volume.
Keeping this in mind, we aim to present a saliency-aware mixup method favorable for transformers.
Specifically, we regard self-attention as the inherent saliency detector, which serves as an efficient means for augmentation.

We accordingly propose \textbf{TokenMixup}, an efficient attention-guided token-level mixup method.
Its objective is to mix intermediate token sets so that the saliency level of a batch is maximized.
To detect salient tokens, we take advantage of the attention map to achieve a $\times15$ speed-up compared to the gradient-based saliency estimator, without compromising model performance.
The batch saliency is maximized by optimally matching instance pairs, accomplished by an algorithm that exactly solves the optimization problem.
% To optimally pair instances for batch saliency maximization, we provide an algorithm that exactly solves the optimization problem.
We also introduce ScoreNet, a simple module that measures the difficulty of an instance.
Based on ScoreNet, TokenMixup is applied selectively with respect to sample difficulty, which can be viewed as a type of curriculum learning method.
To the best of our knowledge, this is the first attempt in the literature to discuss mixup from a curriculum learning perspective.
Furthermore, we present a novel mixup approach that combines intermediate tokens across transformer layers within a single instance.
We name this Vertical TokenMixup (VTM) since it \textit{vertically} mixes tokens to provide rich multi-scale information to the attention layer.

Experiments on image classification datasets imply that our methods are both effective and efficient.
Also, an advantage of TokenMixup is that it can be used in parallel with other mixup methods.
In combination with those methods, we improve our baselines' performance by significant
margins on CIFAR-10, CIFAR-100, and ImageNet-1K.
Especially for CIFAR-100, we achieve a new state-of-the-art performance among transformer-based models that do not use pre-trained models.

Then, our contributions are fourfold:
\begin{itemize}[leftmargin=*]
\item We propose TokenMixup, an efficient attention-guided token-level mixup for transformers. By incorporating the self-attention map, we achieve $\times15$ faster saliency-aware data augmentation compared to gradient-based methods.
% \item We introduce curriculum learning to mixup, which adaptively performs TokenMixup based on the confidence of ScoreNet at an intermediate layer, where TokenMixup is applied.
\item We incorporate curriculum learning into mixup, to adaptively perform TokenMixup in an intermediate layer based on the confidence score of ScoreNet.

% \item Also, \textit{Vertical TokenMixup} is proposed to enable multi-scale TokenMixup within a single instance.
\item We also studied a variant, Vertical TokenMixup, which performs mixup with a single sample and enables multi-scale feature augmentation.

\item We achieve state-of-the-art performance on CIFAR-100 among transformer models trained from scratch. We also gain significant improvements over baseline models for image classification.
\end{itemize}

\section{Related Works}
\label{section:relworks}
\paragraph{Mixup.}
Input Mixup~\cite{zhang2017mixup} is a data augmentation method widely adopted for image classification.
The classification model is trained with a convex combination of input images and labels.
A special case of Input Mixup is CutMix~\cite{yun2019cutmix}, which can be seen as a pixel-wise mixup method with binary masks.
Recent advances in these mixup methods mainly focus on appropriately utilizing saliency information, and on mixing in image feature levels.\\
% \paragraph{Saliency-based Mixup.}
The core motivation of saliency-based mixup is that salient regions should be preserved when mixed, to retain a sufficient amount of information and learn more natural feature representations.
SaliencyMix~\cite{uddin2020saliencymix} adopts various saliency detectors to directly extract salient regions.
Puzzle Mix~\cite{kim2020puzzle} attempts to mix images using saliency information while retaining local statistics.
Co-Mixup~\cite{kim2021co} maximizes gradient-based saliency while encouraging supermodular diversity of the mixed images.
SuperMix~\cite{dabouei2021supermix} takes advantage of supervised signals to mix input images based on saliency.\\
% \paragraph{Feature level Mixup.}
On the other hand, Manifold Mixup~\cite{verma2019manifold} has provided theoretical grounds on the advantages of mixing images in the higher level of features.
Manifold mixing helps the model learn flatter representation and smoother decision boundaries.
Moreover, it is noted that it enhances robustness to adversarial attacks.
Recently, AlignMix~\cite{venkataramanan2021alignmix} was proposed to further align geometric properties of objects by mixing instances in the feature space.
Furthermore, StyleMix~\cite{hong2021stylemix} proposed to manipulate content and style information disparately to generate more abundant and robust samples.

\paragraph{Vision Transformers.}
Originating from the natural language processing field, the Transformer~\cite{vaswani2017attention} model enjoyed tremendous success in numerous computer vision applications, \textit{e.g.}, image classification, object detection~\cite{carion2020end,zhu2020deformable}, and human-object interaction detection~\cite{kim2021hotr,park2022consistency} \textit{et cetera}.
Specifically, image classification is one of the most fundamental computer vision tasks.
Starting from the ViT~\cite{dosovitskiy2020image}, many works have attempted to assimilate the convolutional operation with transformer blocks~\cite{liu2021swin, wang2021pyramid, yuan2021incorporating, heo2021rethinking, han2021transformer, d2021convit}.
Other variants have effectively incorporated the convolutional neural network into the transformer to achieve superior performance~\cite{hassani2021escaping, wu2021cvt, xu2021co, chu2021conditional, chu2021twins}.

\section{Methods}
\label{section:method}
Here, we introduce TokenMixup, a simple but effective mixup method for transformer models.
The goal of our method is to augment intermediate tokens while maximizing the saliency level.
This can be accomplished in three major steps: 1) Sample difficulty assessment, 2) Attention-guided saliency detection \& optimal assignment, 3) Token-level mixup.
The following subsections provide detailed descriptions of the steps.

\begin{figure}[t]
\begin{center}
\includegraphics[width=140mm]{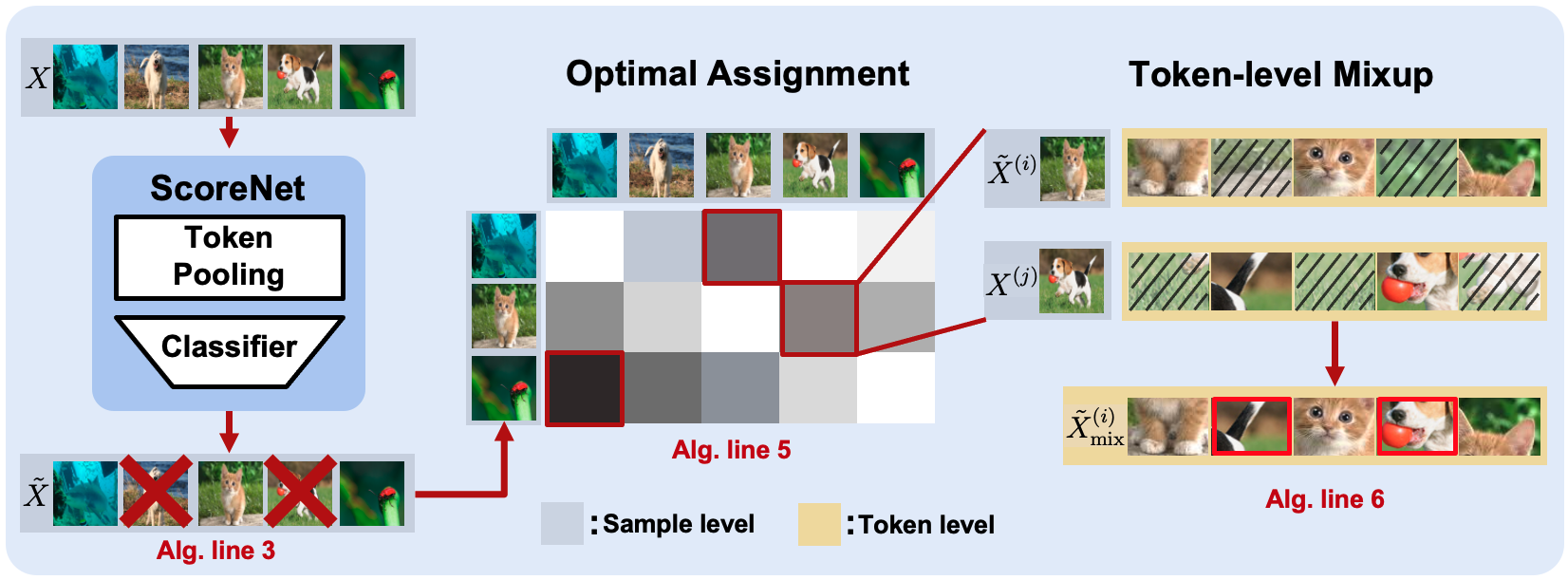}
\end{center}
\vspace{-4mm}
% \caption{\footnotesize \textbf{TokenMixup.} In a transformer layer where TokenMixup is applied, samples in a batch are first filtered with respect to difficulty, resulting in \textit{easy} samples $\tilde{X}$. Then, $\tilde{X}$ is optimally paired with $X$ via hungarian matching, so that overall saliency of the batch is maximized. For a matched pair ($\tilde{X}^{(i)}$, $X^{(j)}$), token indices that reveal saliency difference greater than $\rho$ is replaced, resulting in a new token set with maximum saliency level.}
\caption{\footnotesize \textbf{TokenMixup.} Batch samples $X$ are first filtered with respect to difficulty scores evaluated by ScoreNet in $\mathcal{F}$, resulting in \textit{easy} samples $\tilde{X}$. 
Then, images in $\tilde{X}$ are optimally paired with samples in $X$ via Hungarian matching, so that the overall saliency of the mini-batch is maximized after mixup. 
For a matched pair ($\tilde{X}^{(i)}$, $X^{(j)}$), tokens of $\tilde{X}^{(i)}$ with sufficiently lower saliency levels than the tokens of $X^{(j)}$ are replaced, resulting in a new token set $\tilde{X}^{(i)}_{\textrm{mix}}$ containing a greater saliency level.}
\label{fig:htm}
\end{figure}

\subsection{Sample Difficulty Assessment}
\label{sec:scorenet}
We first assess the difficulty of the input to adaptively decide whether to apply augmentation.
To evaluate sample difficulty, a measuring function $\mathcal{F}$ needs to be defined.
% which evaluates the difficulty of each sample.
We utilize a parameterized module, ScoreNet, which is a simple MLP that predicts the target value $Y^{(i)}$ based on the intermediate tokens $X^{(i)}$, where $i$ is the index of the sample in a batch \textcolor{black}{(See the supplement for figure)}.
Given the ScoreNet output, the difficulty score is evaluated with the prediction loss computed as
\begin{equation}
\label{eq:difficulty}
    \mathcal{F}(X^{(i)}, Y^{(i)}) = \textrm{CrossEntropy}(\textrm{ScoreNet}(X^{(i)}), Y^{(i)}).
\end{equation}
If the score is greater than a threshold $\tau$, the sample is deemed sufficiently hard and no mixup is performed.
On the other hand, the set of \textit{easy} samples with $\mathcal{F}(X^{(i)}, Y^{(i)}) < \tau$ denoted as $\tilde{X}$ will be augmented by Token-level mixup after pairing with other samples in a mini-batch, as illustrated in Figure~\ref{fig:htm}.
Note that ScoreNet is trained simultaneously with the main model and training ScoreNet to assess sample difficulty at the intermediate layer can be viewed as an auxiliary task. 
% That is, by defining the loss function as $\mathcal{L}_{cls} + \lambda \cdot \mathcal{F}(X^{(i)})$, ScoreNet also serves as an auxiliary loss function.

% To adaptively apply augmentation based on sample difficulty, a function $\mathcal{F}$ needs to be defined which evaluates the difficulty of each sample.
% We use a parameterized module, ScoreNet, as the difficulty measurer.
% ScoreNet is a simple MLP module that attempts to predict the target value based on the intermediate tokens (See Appendix A.1 for figure).
% The prediction of ScoreNet is used to compute the cross entropy loss, which we adopt as the difficulty score.
% Then, function $\mathcal{F}$ can be written as
% \begin{equation}
%     \mathcal{F}(X^{(i)}) = \textrm{CrossEntropy}(\phi(\textrm{ScoreNet}(X^{(i)}), Y^{(i)})),
% \end{equation}
% where $X^{(i)}, Y^{(i)}$ are the $i^{th}$ batch instance, and $\phi$ refers to the softmax function.
% If the score is greater than threshold $\tau$, the sample is deemed sufficiently hard and no mixup is performed.
% On the other hand, we denote the samples with $\mathcal{F}(X^{(i)}) < \tau$ as $\tilde{X}$, as found in Figure~\ref{fig:htm}.
% In addition, note that ScoreNet is trained simultaneously with the main model, whose loss is $\mathcal{L}_{cls}$.
% That is, by defining the loss function as $\mathcal{L}_{cls} + \lambda \cdot \mathcal{F}(X^{(i)})$, ScoreNet also serves as an auxiliary loss function.

\subsection{Attention-guided Saliency Detection \& Optimal Assignment}
Several works~\cite{heo2021rethinking, rao2021dynamicvit, roh2021sparse, wang2021pnp} have discussed the imbalanced information of tokens.
Due to this imbalance, random token mixing would potentially cause significant information loss and meaningless token replacements (See the supplement for relevant analysis).
Thus, we aim to mix tokens based on saliency.

\vspace{-2mm}
\paragraph{Attention-guided saliency detection.} 
Instead of the computationally heavy gradient-based saliency detectors~\cite{kim2020puzzle, kim2021co}, we take advantage of the attention map, an inherent saliency approximator within transformer modules.
Specifically, we can infer the saliency of the tokens in the $i^{th}$ layer via Attention Rollout~\cite{abnar2020quantifying}, which computes the attention imposed on the tokens from layer $i$ to $i + \ell$ ($\ell \geq 0$) as
\begin{equation}
\label{eq:rollout}
    A = \Phi^{(i)} \cdot \Phi^{(i+1)} \cdot \cdot \cdot \Phi^{(i+\ell)},
\end{equation}
such that $\Phi^{(i)} = \frac{1}{H}\sum_{h=1}^H \Phi^{(i)}_h$ where $H$ is the number of heads in the multi-head attention layer, and $\Phi^{(i)}_h$ is the $h^{th}$ attention head of layer $i$.
In order to retrieve $A$, the samples need to be propagated for the subsequent $\ell$ layers with stop-gradient.
Since each additional forward propagation is an overhead, we approximate Attention Rollout by setting $\ell = 0$.
That is, we only use the attention map of the following layer as the saliency estimator. 
We find this a sufficient approximation of the full Attention Rollout.
See section~\ref{sec:sal_sound} for analysis on the soundness of the approximated saliency map.
Then, the saliency score $S_t$ can be computed as
\begin{equation}
\label{eq:saliencyscore}
    S_t = \displaystyle \frac{1}{n} \sum_{i=1}^n A_{i, t},
\end{equation}
where $t = 1, 2, \ldots, n$ denotes the token index.

\paragraph{Optimal assignment.} 
Based on the estimated saliency of tokens, we aim to maximize the overall saliency level by optimally assigning a different mixup target for each instance.
So, we first define $ P_{i,j} \in \rm I\!R^{n}$ as the saliency difference between a random instance pair $(i, j)$, which is computed as
\begin{equation}
\label{eq:definition}
    P_{i,j} \overset{\triangle}{=} S_j - S_i,
\end{equation}
where $i, j = 1, 2, \ldots, b$ and $S_i \in \rm I\!R^n$ is the token saliency map from the $i^{th}$ instance.
Then, we represent our objective as an optimization problem written as
\begin{equation}
\label{eq:htm_objective}
    \max_{\sigma \in \mathcal{M}} \: \max_{r_i \in \rm I\!B^n} \: \displaystyle \sum _{i=1}^b (P_{i, \sigma(i)}-\rho)^\top  r_i ,
\end{equation}
where $\mathcal{M}$ is the set of all possible batch permutations of the $b$ instances, and $\sigma \in \mathcal{M}$ refers to an arbitrary permutation.
$\rho$ is the threshold hyperparameter that controls the minimum saliency gain required for a token to be mixed, and $r_i$ is a binary decision vector for the $i^{th}$ instance. 
\textit{i.e.}, token $t$ of $i$ is replaced with token $t$ of $\sigma(i)$ when $r_{i,t} = 1$, and preserved when $r_{i,t} = 0 \:\: (t = 1, 2, \ldots, n)$.

The above optimization problem can be exactly solved by utilizing the Hungarian Matching algorithm~\cite{munkres1957algorithms}.
% The score matrix, $C_{i,j} \in \rm I\!R^{n \times n}$, for the matching algorithm is computed as
The matching algorithm requires the score matrix, $C \in \rm I\!R^{b \times b}$, which is computed as
\begin{equation}
\label{eq:matchscore}
    C_{i,j} = \displaystyle \sum_{t=1}^n \textrm{max}(P_{i,j}^{(t)} - \rho,\: 0),
\end{equation}
where each item in $C_{i,j}$ refers to the maximum saliency gain resulting from mixing $\tilde{X}^{(i)}$ and $X^{(j)}$.
The actual maximum gain will be reached when $\rho = 0$, but we set a positive threshold to control the minimum saliency gain required for each token to be replaced.
Then, by applying the Hungarian algorithm, we can find the optimal batch permutation $\sigma^*$ such that
\begin{equation}
    \sigma^* = \displaystyle \underset{\sigma}{\arg\max} \sum_{i=1}^n C_{i,\sigma(i)}.
\end{equation}
Then, by mixing $\tilde{X}^{(i)}$ and $X^{(\sigma^*(i))}$, our objective is optimized.

\begin{algorithm}[t]
\caption{TokenMixup}
\label{alg:htm_v2}
\textbf{Input:} $X \in \rm I\!R^{b \times n \times d}$, $Y \in \rm I\!R^{b \times c}$, $\tau$, $\rho$\\
\textbf{Output:} $X \in \rm I\!R^{b \times n \times d}$, $Y \in \rm I\!R^{b \times c}$
\begin{algorithmic}[1]
    \State $U^{(i)} \gets \textrm{evaluate difficulty of } X^{(i)} \:\:\:\:$ s.t.  $i = 1, 2, \ldots, b$ \hfill $\triangleright$ Eq.~\eqref{eq:difficulty}
    \State $S_t^{(i)} \gets \textrm{evaluate saliency of each token with $\ell$-step attention rollout}$ \hfill $\triangleright$ Eq.~\eqref{eq:saliencyscore}
    \State $\tilde{X}^{(i)}, \tilde{Y}^{(i)}, \tilde{S}^{(i)} \gets $ select easy samples w.r.t $U$ and $\tau$  \:\:\:\: s.t.  $i = 1, 2, \ldots, b'$
    \State $C_{ij} \gets \sum_{t} \textrm{max}(S_t ^{(j)} - \tilde{S}_t^{(i)} - \rho,\:  0) \:\:\:\:$ s.t. $i = 1, 2, \ldots, b'$ and $j = 1, 2, \ldots, b$ \hfill $\triangleright$ Eq.~\eqref{eq:matchscore}
    \State $\sigma(m) \gets \textrm{HungarianMatching}(C_{ij})\:\:\:\:$ s.t. $m = 1, 2, \ldots, b'$ and $\sigma(m) \in \{1, 2, \ldots, b\}$
    \State $\tilde{X}^{(i)}_{\textrm{mix}} \gets$ Mix-token($\tilde{X}^{(i)}, X^{(\sigma(i))}; \rho)$ \:\:\:\: s.t. $i = 1, 2, \ldots, b'$ \hfill $\triangleright$ Eq.~\eqref{eq:mix_x}
    \State $\tilde{Y}^{(i)}_{\textrm{mix}} \gets$ Relabel($\tilde{Y}^{(i)}, Y^{(\sigma(i))})$ \hfill $\triangleright$ Eq.~\eqref{eq:mix_y}
    \State restore $\tilde{X}^{(i)}_{\textrm{mix}}, \tilde{Y}^{(i)}_{\textrm{mix}}$ to $X$, $Y$ \:\:\:\: s.t. $i = 1, 2, \ldots, b'$
    \State \textbf{return } $X$, $Y$
\end{algorithmic}
\end{algorithm}

By incorporating sample difficulty assessment (section~\ref{sec:scorenet}) into~\eqref{eq:htm_objective}, our final objective is written as
\begin{equation}
\label{eq:htm_objective2}
\begin{aligned}
    \max_{\sigma \in \mathcal{M}} \: & \max_{r_i \in {\rm I\!B^n}} \: \displaystyle \sum _{i=1}^b \mathbb{1}_{\{\mathcal{F}(X^{(i)}, Y^{(i)}) < \tau \}}(P_{i, \sigma(i)}-\rho)^\top  r_i \\ & \textrm{s.t.} \:\: \mathbf{1}^\top r_i \leq n \times \mathbb{1}_{\{\mathcal{F}(X^{(i)}, Y^{(i)}) < \tau\}},
\end{aligned}
\end{equation}
where $\mathcal{F}$ is the difficulty measurer, and $\tau$ is the difficulty threshold.
$\mathbb{1}$ denotes the indicator function, while $\mathbf{1}$ is a one vector.
If the sample is hard, \textit{i.e.}, $\mathcal{F}(X^{(i)}, Y^{(i)}) \geq \tau$, the constraint becomes $\mathbf{1}^\top r_i \leq 0$, enforcing all elements of $r_i$ to be 0. 
In that case, none of the tokens will be mixed.
On the other hand, if the sample is easy, \textit{i.e.}, $\mathcal{F}(X^{(i)}, Y^{(i)}) < \tau$, the constraint is $\mathbf{1}^\top r_i \leq n$ which is satisfied by any random $r_i$ vector.
Then, the objective becomes identical to~\eqref{eq:htm_objective}.

\subsection{Token-level Mixup}
Now, the remaining question is \textit{how} the tokens of $\tilde{X}^{(i)}$ and $X^{(\sigma^*(i))}$ are going to be mixed.
As defined in objective~\eqref{eq:htm_objective}, we use a hard replacement of tokens with respect to decision vector $r_i$.
Then, mixing is performed by replacing token index $t$ from the original instance $i$, with token $t$ from the paired instance $j=\sigma^*(i)$, so that $P_{i,j}^{(t)} > \rho$ is satisfied.
In other words, only the token indices that increase saliency level greater than $\rho$ are mixed.
This is accomplished by defining an appropriate binary mask $M_t \in \rm I\!B^n \; (t=1,2,\ldots,n)$ containing
\begin{equation}
\label{eq:mask}
    M_t = 
    \begin{cases}
    \hspace{2mm}0  & ,\:\: S_t^{(j)} - \tilde{S}_t^{(i)} > \rho \\
    \hspace{2mm}1  & ,\:\: \textrm{otherwise}
    \end{cases}
\end{equation}
where $\tilde{S}_t^{(i)}$ is the saliency score of $\tilde{X}^{(i)}_t$.
Then, the mask is employed as 
\begin{equation}
\label{eq:mix_x}
    \tilde{X}^{(i)}_{\textrm{mix}} = M \odot \tilde{X}^{(i)} + (1-M) \odot X^{(j)},
\end{equation}
where operator $\odot$ denotes element-wise multiplication and $\tilde{X}^{(i)}$, $X^{(j)}$ refer to the paired instances.
We also reassign labels based on the replaced tokens' saliency score.
The new mixed label is computed as
\begin{equation}
\label{eq:mix_y}
    \tilde{Y}_{\textrm{mix}}^{(i)} = \frac{\sum_{t=1}^n M_t \cdot \tilde{S}_t^{(i)}}{\sum_{t=1}^n M_t \cdot \tilde{S}_t^{(i)} + (1-M_t) \cdot S_t^{(j)}} \cdot \tilde{Y}^{(i)} + \frac{\sum_{t=1}^n (1-M_t) \cdot S_t^{(j)}}{\sum_{t=1}^n M_t \cdot \tilde{S}_t^{(i)} + (1-M_t) \cdot S_t^{(j)}} \cdot Y^{(j)}.
\end{equation}

Consequently, relatively unimportant tokens will be replaced with salient tokens from another instance, and its label will be adjusted with respect to the change in overall saliency level. 
See Figure~\ref{fig:htm} for visualization, and Algorithm~\ref{alg:htm_v2} for overall computation process. 
More detailed pseudocode is provided in the supplement.
\subsection{Vertical TokenMixup}

\begin{figure}[t]
\begin{center}
\includegraphics[width=130mm]{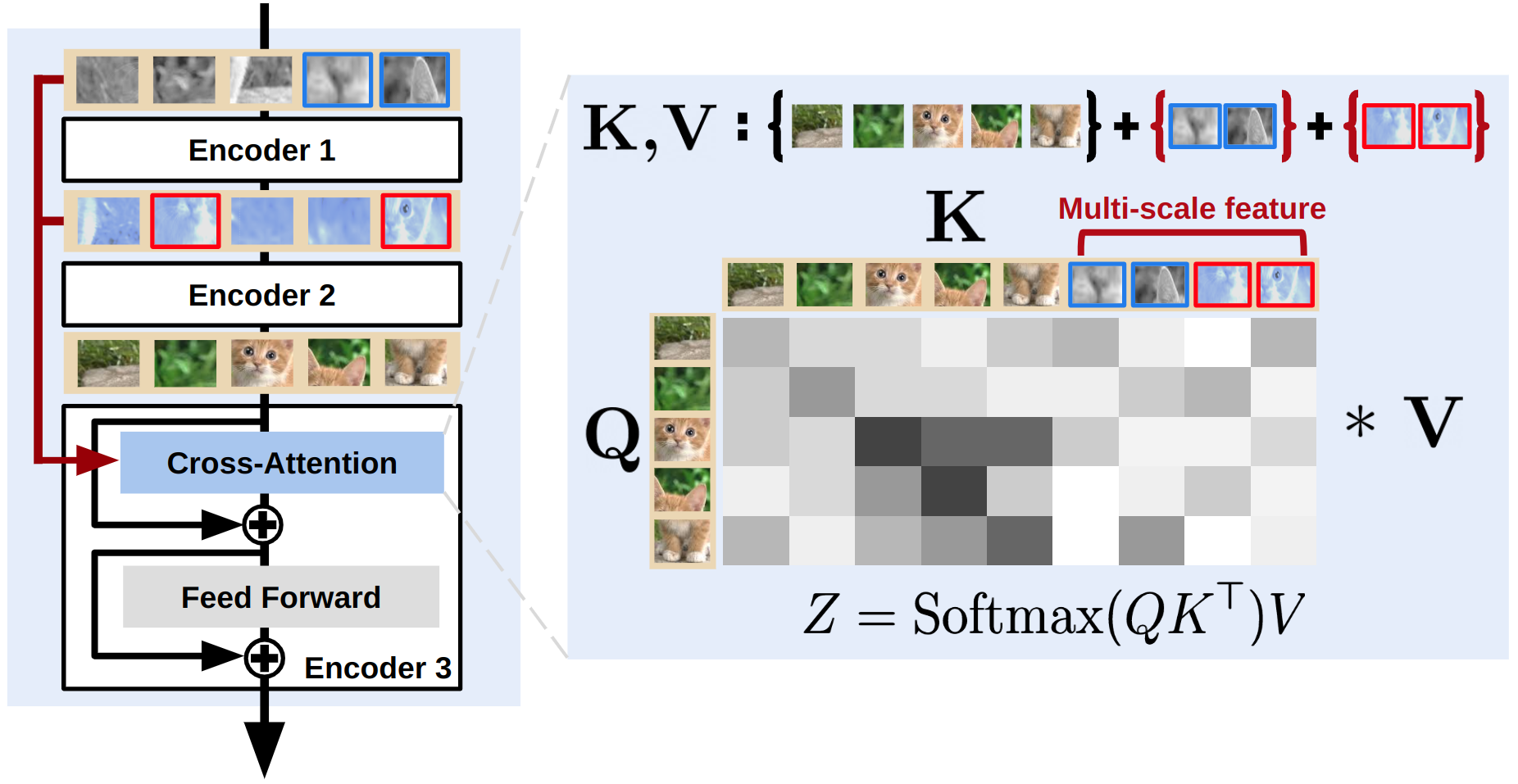}
\end{center}
\vspace{-4mm}
\caption{\footnotesize \textbf{Vertical TokenMixup}. By applying VTM, the $\kappa$ most salient tokens from each previous layer are brought up to form the key and value tokens. By utilizing a cross-attention mechanism, the dimension is preserved while being able to aggregate multi-scale information for token augmentation. The figure demonstrates the case when VTM is applied to layer 3. }
\label{fig:vtm}
\end{figure}
Mixing multiple \textit{instances} is not the only way to mixup.
We also present a simple variant of TokenMixup, which 
provides token-level augmentation within a single sample by utilizing the tokens from previous layers.
% increases the saliency level within a single instance by utilizing the tokens from previous layers.
Similar to \eqref{eq:htm_objective}, we can define a new objective for Vertical TokenMixup as
\begin{equation}
    \max_{l \in \mathcal{L}} \: \max_{r_i \in \rm I\!B^n} \: \displaystyle \sum _{i=1}^b (P_{i, l(i)}-\rho)^\top \cdot r_i ,
\end{equation}
where $\mathcal{L}$ is the set of previous layer indices, and $l(i)$ returns an arbitrary layer index for the $i^{th}$ instance.
This objective can be optimized by a similar scheme as in Algorithm \ref{alg:htm_v2}.
However, considering that the tokens from each layer with an identical index tend to contain similar information, saliency difference matrix $P_{i,l(i)} = S_{l(i)} - S_i$ (as in~\eqref{eq:definition}) will most likely be constant across indices.
This may lead to meaningless mixup without much increase in saliency level.
Therefore, we take a different approach to mix tokens \textit{vertically}.

The simplest method would be to concatenate all tokens from previous layers and apply self-attention.
% But this may induce quadratic complexity, and the resulting output dimension will chagn
But due to the quadratic complexity of self-attention, naive concatenation may induce excessive computation.
To reduce overhead, we selectively pool the $\kappa$ most salient tokens from each previous layer and concatenate them to the original token set.
Also, to preserve input dimension, we adopt the cross-attention mechanism.
If $X_1$ is the original token set and $X_2$ is the concatenated token set, vertical cross-attention is expressed as
\begin{equation}
    Z = \textrm{Softmax}(X_1 W_q  (X_2 W_k)^\top) \hspace{1mm} X_2 W_v,
\end{equation}
where $W_q, W_k, W_v$ denote the projection layers for query, key, and value, respectively.
% By setting $\kappa \ll n$, where $n$ is the number of tokens per layer, complexity is reduced from $\mathcal{O}(|\mathcal{L}|^2\cdot n^2)$ to $\mathcal{O}(n^2 + (|\mathcal{L}|-1)\cdot\kappa \cdot n)$.
See Figure~\ref{fig:vtm} for visualization, and the supplement for pseudocode.
\subsection{Discussions}

% \paragraph{Difference with Co-Mixup.}
% TokenMixup can be viewed as a variation of Co-Mixup~\cite{kim2021co} applied to manifold-level transformer tokens.
% Both share an objective of finding the optimal mixup pairs within a mini-batch that maximize overall saliency.
% However, the key difference is how saliency map is derived.
% Due to huge parameter sets of transformers, gradient-based saliency detection can be a computational bottleneck.
% We alleviate this clutter by approximating the saliency map with the transformer attention map.
% % This requires only one layer forward pass on a self-attention layer without any backward propagation.
% This requires only one-step forward pass on a self-attention layer, compared to the full forward pass followed by a backward pass of gradient-based methods.
% We show in section~\ref{sec:sal_sound} that this approximation is sufficient for saliency detection.
% Another interesting property of transformers is that each intermediate token is a global aggregation of features.
% This relieves the constraint of local smoothness, which is one of the major optimization objectives in Co-Mixup.
% % Thus, in the case of TokenMixup, a simple drop and insert mechanism can be applied to tokens with less concerns for local statistics.
% Thus, we may simply drop and insert tokens with less concerns for local statistics.
% \vspace{-2mm}
\paragraph{TokenMixup as Curriculum Learning.}
% A novelty of TokenMixup is that it is a type of predefined curriculum learning~\cite{wang2021survey} method.
% One component of predefined curriculum learning is the difficulty measurer, whose role is taken by the ScoreNet.
% The performance of the ScoreNet is a lower bound of the entire transformer model, as it attempts to classify with intermediate features.
% Thus, its performance can be a direct proxy for the input instance difficulty, and TokenMixup can be applied only when the input is expected to be easy.
% This operation is especially helpful when TokenMixup is applied synchronously with other data augmentation methods, \textit{e.g.} input mixup, cutmix, or affine transformations.
% That is, TokenMixup can control the augmentation intensity by refraining from mixing already augmented samples.
% Furthermore, most samples will have low accuracy in the earlier stage of training, leading to lower mixup frequency.
% If the model converges, on the other hand, most samples will become \textit{easy}, and further augmentation will be applied to make the task more challenging.
% In this sense, TokenMixup is a training scheduler that automatically schedules the difficulty of an input, which is the other key component of predefined curriculum learning.
A general framework for curriculum learning consists of two main components: Difficulty Measurer and Training Scheduler~\cite{wang2021survey}.
TokenMixup satisfies these conditions, as ScoreNet measures the difficulty of each input, and augmenting the tokens based on input difficulty naturally schedules training.
In the early training phase where the model parameter is not optimized, most instances will be evaluated \textit{difficult} and no TokenMixup is performed.
As the model converges, on the other hand, many samples will become \textit{easy}, triggering augmentation to make the task more challenging overall.
Also, TokenMixup is applied discriminatively based on individual sample difficulty which enables more intricate curriculum scheduling.
See section~\ref{sec:analysis_learn} for empirical demonstration.
These properties will be especially useful when other mixup methods~\cite{zhang2017mixup, yun2019cutmix, kim2020puzzle, kim2021co} are used in parallel.
For instance, if input mixup is already applied, further augmentation may be unnecessary.
In such a case, ScoreNet will regard the sample sufficiently difficult, and no further mixup will be performed.

\section{Experiments}
\label{section:experiments}

\subsection{Preliminaries}
\vspace{-2mm}
\paragraph{Baseline.}
% TokenMixup is a general augmentation method for transformer models.
In experiments on CIFAR~\cite{krizhevsky2009learning}, we used Compact Convolution Transformer (CCT)~\cite{hassani2021escaping} as our baseline.
% CCT is a vision transformer model that adequately adopts the convolutional neural network along with its sequence pooling technique.
% CCT achieves state-of-the-art performance on small datasets like MNIST and CIFAR~\cite{krizhevsky2009learning} with relatively few parameters and without any pretrained weights.
For ImageNet-1k~\cite{ILSVRC15} experiments, we used the vanilla ViT-B/16~\cite{dosovitskiy2020image} as baseline.
Then, our augmentation methods are evaluated on three representative image classification datasets: CIFAR-10, CIFAR-100, and ImageNet-1K.

\vspace{-2mm}
\paragraph{Experimental setup.}
We evaluate TokenMixup and its variant, Vertical TokenMixup (VTM), on CIFAR and ImageNet with different settings.
To avoid confusion, we denote the original TokenMixup as Horizontal TokenMixup (HTM) in the following sections.
For CIFAR experiments, we adopt the 1500-epoch version of CCT.
We modified the learning rate scheduler and positional embedding type to achieve better performance than original papers (denoted $*$ in Table~\ref{tab:cifar}).
% The boosted model is denoted with superscript $*$ in Table~\ref{tab:cifar}.
Other experiment settings follow~\cite{hassani2021escaping}, and all experiments on CIFAR datasets were conducted on a single RTX A6000 GPU.
For ImageNet-1K experiments, we used the ViT-B/16 model pre-trained and fine-tuned on ImageNet-21k and ImageNet-1k, where we took advantage of the officially released pre-trained weights.
Experiments for Horizontal TokenMixup were conducted on a single NVIDIA A100 GPU, and 4 RTX 3090 GPUs were used in parallel for Vertical TokenMixup.
Other experiment settings are reported in the supplement.

\vspace{-2mm}

\subsection{Experimental Results}
\vspace{1mm}

\begin{table}[t]
    \centering
    \setlength{\tabcolsep}{3.5pt}
    \caption{\footnotesize \textbf{Experimental results on CIFAR.} Experiment results with TokenMixup on CCT are compared with state-of-the-art models that do not use pretrained models for initialization. Top-1 validation accuracy is reported for each model, and baseline models denoted with * reports retrained performance. Also, the number in the parenthesis indicates the number of epochs used for training, and the parameter numbers are retrieved from the CIFAR-100 models. Best performances are highlighted in yellow.}
    \begin{adjustbox}{width=0.9\textwidth}
    \begin{tabular}{l@{\hskip 0.2in} |@{\hskip 0.1in} c c@{\hskip 0.1in} |@{\hskip 0.1in} c@{\hskip 0.1in} c@{\hskip 0.1in}}
        \toprule
        \textbf{Models} & \textbf{\# Params} & \textbf{MACs} & \textbf{CIFAR-10} & \textbf{CIFAR-100}  \\
        \midrule
        \midrule
        \multicolumn{5}{l}{\textit{Convolutional Network based}} \\ 
        \midrule
        ResNet18 & 11.18 M & 0.04 G & 90.27 & 63.41 \\
        % ResNet34 & 21.29 M & 0.08 G & 90.51 & 64.52 \\
        ResNet50 & 23.53 M & 0.08 G & 90.60 & 61.68 \\
        ResNet1001-v2~\cite{he2016identity} & 10.33 M & 1.55 G & 95.08 & 77.29 \\
        MobileNetV2/2.0~\cite{sandler2018mobilenetv2} & 8.72 M & 0.02 G & 91.02  & 67.44 \\
        WRN-28-10~\cite{zagoruyko2016wide} & 36.5 M & - & 96.00 & 80.75 \\
        WRN-40-4~\cite{zagoruyko2016wide} & 8.9 M & - & 95.47 & 78.82 \\
        ResNeXt-29-8×64d~\cite{xie2017aggregated} & 34.4 M & - & 96.35 & 82.23 \\
        ResNeXt-29-16×64d~\cite{xie2017aggregated} & 68.1 M & - & 96.42 & 82.69 \\
        \midrule
        \multicolumn{5}{l}{\textit{Transformer based}} \\ 
        \midrule
        % ViT-12/16 & 85.63 M & 0.43 G & 76.42 & 46.61 \\
        ViT-Lite-6/4~\cite{hassani2021escaping}  & 3.19 M & 0.22 G & 93.98 & 73.33 \\
        ViT-Lite-7/4~\cite{hassani2021escaping}  & 3.72 M & 0.26 G & 93.57 & 73.94 \\
        NesT-T~\cite{zhang2022nested} & 17.0 M & - & 96.04 & 78.69 \\
        NesT-B~\cite{zhang2022nested} & 68.0 M & - & 97.20 & 82.56 \\
        CVT-6/4~\cite{hassani2021escaping} & 3.19 M & 0.22 G & 93.60 & 74.23  \\
        CCT-7/3x1(1500)~\cite{hassani2021escaping}  & 3.76 M & 0.95 G & 97.48 & 82.72 \\
        CCT-7/3x1(1500) $^{*}$  & 3.78 M & 0.95 G & 97.48 & 82.87  \\
        CCT-7/3x1(1500) + \textbf{Horizontal TM (ours)} & 3.81 M  & 0.95 G & \textbf{97.57}  & \textbf{83.56}  \\
        CCT-7/3x1(1500) + \textbf{Vertical TM (ours)} & 3.78 M & 0.95 G & \cellcolor{yellow!25}\textbf{97.78} & \textbf{83.54} \\ 
        CCT-7/3x1(1500) + \textbf{HTM + VTM (ours)} & 3.81 M & 0.95 G & \textbf{97.75} & \cellcolor{yellow!25}\textbf{83.57} \\ 
        \bottomrule
    \end{tabular}
    \end{adjustbox}
    \label{tab:cifar}
\end{table}
\vspace{-4mm}
\begin{table}[t]
    \centering
    \setlength{\tabcolsep}{3.5pt}
    \caption{ \footnotesize \textbf{Experimental results on ImageNet.} Experiment results with TokenMixup on ViT are compared. We report Top-1 and Top-5 validation accuracy on ImageNet-1k. For the baseline model, we used ViT-B/16 ($224\times224$). $^\dagger$ denotes fine-tuned performance officially reported by the authors of ViT.}
    \begin{adjustbox}{width=0.9\textwidth}
    \begin{tabular}{l@{\hskip 0.3in} |@{\hskip 0.1in} c c@{\hskip 0.1in} |@{\hskip 0.1in} c@{\hskip 0.1in} c@{\hskip 0.1in}}
        \toprule
        \textbf{Models} & \textbf{\# Params} & \textbf{MACs} & \textbf{Top-1 Acc.} & \textbf{Top-5 Acc.}  \\
        \midrule
        \midrule
        \multicolumn{5}{l}{\textit{Convolutional Network based}} \\ 
        \midrule
        ResNet50~\cite{he2016deep} & 25.5 M & 4.3 G & 76.20 & - \\
        ResNet152~\cite{he2016deep} & 60.19 M & 11.58 G & 78.57 & - \\
        DenseNet-264~\cite{huang2017densely} & - & - & 79.20 & 94.71 \\
        WRN-50-2-bottleneck~\cite{zagoruyko2016wide} & 68.9 M & - & 78.10 & 93.97 \\
        ReGNetY-4G~\cite{radosavovic2020designing} & 21 M & 2.0 G & 80.0 & - \\
        ResNeXt-50-2×40d~\cite{xie2017aggregated} & - & - & 77.00 & - \\
        ResNeXt-101-64x4d~\cite{xie2017aggregated} & - & - & 79.60 & 94.70 \\
        \midrule
        \multicolumn{5}{l}{\textit{Transformer based}} \\ 
        \midrule
        ViT-B/16-224$^{\dagger}$ ~\cite{dosovitskiy2020image}  & 86.6 M & 16.9 G  & 81.2 & - \\
        ViT-B/16-224 + \textbf{Horizontal TM (ours)} & 87.3 M & 16.9 G &  \cellcolor{yellow!25}\textbf{82.37} & \textbf{96.29} \\
        ViT-B/16-224 + \textbf{Vertical TM (ours)} & 86.6 M & 16.9 G & \textbf{82.30} & \cellcolor{yellow!25}\textbf{96.33} \\ % 82.31 => 82.29
        ViT-B/16-224 + \textbf{HTM + VTM (ours)} & 86.6 M & 16.9 G & \textbf{82.32} & \textbf{96.27} \\ % 82.31 => 82.29
        \bottomrule
    \end{tabular}
    \end{adjustbox}
    \label{tab:imagenet}
\end{table}

First, we demonstrate the image classification performance on CIFAR-10 and CIFAR-100 when TokenMixup is applied to CCT.
From Table~\ref{tab:cifar}, we can see that applying Horizontal TokenMixup and Vertical TokenMixup simultaneously improves its strict CIFAR-100 baseline, by a significant margin of 0.70\%.
HTM and VTM each surpasses the baseline performance by a margin of 0.69\% and 0.67\% as well.
Also, we outperform the previous state-of-the-art model (among from-scratch models) by a huge margin of 0.85\%.
% This is the state-of-the-art performance on CIFAR-100 among transformers models that do not use pretrained weights at initialization.
% We outperform the previous state-of-the-art accuracy by a huge margin of 0.83\%.\\
In the case of CIFAR-10, we achieved higher accuracy of 97.57\% and 97.78\% with HTM and VTM, respectively.
% In the case of CIFAR-10, we achieved higher accuracy of 97.65\% and 97.84\%, respectively with HTM and VTM, compared to their baseline settings.
Considering the highly saturated performance on this dataset, a 0.30\% improvement over the baseline is not trivial.\\
Performance on ImageNet-1K is evaluated on the vanilla ViT~\cite{dosovitskiy2020image} model.
The officially reported Top-1 accuracy is 81.2\%, which is achieved by using weights pre-trained on ImageNet-21K and fine-tuning on ImageNet-1K.
By fine-tuning the model further by employing our augmentation methods, we achieve Top-1 accuracy of 82.37\% with Horizontal TokenMixup, 82.30\% with Vertical TokenMixup, and 82.32\% by using both.
Also note that HTM marginally increases parameter size by only 0.7M, while VTM does not require any additional parameters.
See the supplement for further experiments on other baselines.
% \input{04_Experiments/cifar}
% \input{04_Experiments/imagenet}
% \vspace{-2mm}
\section{Analysis}
\label{section:analysis}
% \vspace{-3mm}
Here, we analyze TokenMixup to answer the following research questions: 
\textbf{Q1}. Is the Attention  an appropriate saliency approximation? 
\textbf{Q2}. What are the effects of each component in TokenMixup? 
\textbf{Q3}. What properties does TokenMixup have?

\begin{figure}[t]
\begin{center}
\includegraphics[width=\textwidth]{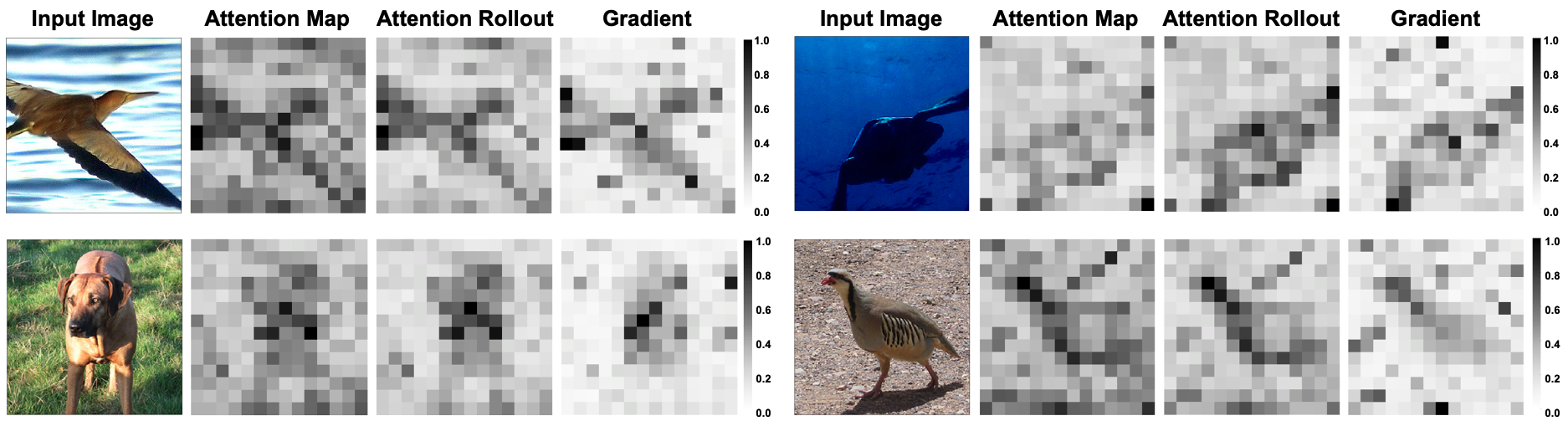}
\end{center}
\vspace{-3mm}
\caption{\footnotesize \textbf{Qualitative comparison of saliency detectors.} The saliency map derived for each method is provided. We can observe that the methods render similar outputs. Note, the values are minmax scaled to [0,1].}
\label{fig:rollout}
% \vspace{-3mm}
\end{figure}
\begin{table}

\centering
\caption{\footnotesize \textbf{Saliency detector comparison.} Our method (HTM) achieves high accuracy while being $\times15$ as fast as the gradient method. Also, we observe only $\times3$ latency compared to the random baseline.}
\label{tab:saliency_map}
\begin{adjustbox}{width=0.8\textwidth}
\begin{tabular}{c|ccc}
\hline
\rule{0pt}{1.0\normalbaselineskip}
\cellcolor{gray!10}Saliency type & \cellcolor{gray!10}\textbf{Gradient-based} & \cellcolor{gray!10}\textbf{Attention-based} &\cellcolor{gray!10} \textbf{Random baseline} 
\rule{0pt}{1.0\normalbaselineskip}\\
\hline
\rule{0pt}{1.0\normalbaselineskip}
Avg. Latency (ms) & 236 {\footnotesize (\textcolor{red}{$\times$14.8})} & 16 {\footnotesize ($\times$1.0)} & 5 {\footnotesize (\textcolor{blue}{$\times$0.3})} \rule{0pt}{1.0\normalbaselineskip}\\
\rule{0pt}{1.0\normalbaselineskip}
CIFAR-100 Acc. & 83.28 &  \textbf{83.56} & 83.05
\rule{0pt}{1.0\normalbaselineskip}\\ % 83.18 => 83.28
\hline
\end{tabular}
\end{adjustbox}
\end{table}

%  \cellcolor{yellow!25} 
% \vspace{-3mm}
\subsection{Soundness of Saliency Estimation}
\label{sec:sal_sound}
\vspace{-2mm}
In TokenMixup, we estimate the saliency using the attention map from the subsequent layer.
The soundness of this approximation would be one of the main questions regarding saliency detection.
So, we provide relevant analyses on the appropriateness of our saliency approximation method to answer the research question \textbf{Q1}.
\paragraph{Qualitative examples.}
\vspace{-3mm}
Here, we provide a qualitative analysis on the performance of each saliency detector.
In Figure~\ref{fig:rollout}, we can observe that there is no significant difference between the full attention rollout and the 1-step approximation, \textit{i.e.}, the subsequent attention map.
Slight discrepancy in the sharpness of the distribution can be observed, but it does not greatly affect our method that utilizes saliency threshold $\rho$.
Moreover, attention-based methods seem to be more accurate than the gradient-based detector in certain cases.
See the supplement for additional qualitative examples, and also the quantified comparison on the sharpness measures of the two methods.
\paragraph{Efficiency comparison.}

\begin{table}[t]
    \centering
    \setlength{\tabcolsep}{3.5pt}
    \caption{\footnotesize\textbf{Ablation study on CIFAR-100.} Performance on CIFAR-100 is evaluated by ablating each component one-by-one. We observe that optimal assignment and label reassignment plays a significant role.}
    \vspace{-3mm}
        \begin{adjustbox}{width=0.7\textwidth}
    \footnotesize
    \begin{tabular}[t]{l | c}
    \toprule
    \textbf{Ablated Component} & \textbf{CIFAR-100 Accuracy} \\
    \midrule
    CCT-7/3x1 + HTM & \textbf{83.56} (- 0.00) \\
    \hspace{4mm}(--) Sample difficulty assessment (ScoreNet) & 82.85 (\textcolor{blue}{- 0.71}) \\
    \hspace{4mm}(--) Optimal assignment & 82.61 (\textcolor{blue}{- 0.95}) \\
    \hspace{4mm}(--) Token-level mixup & 83.05 (\textcolor{blue}{- 0.51}) \\
    \hspace{4mm}(--) Saliency-based label reassignment \hspace{10mm} & 82.62 (\textcolor{blue}{- 0.94}) \\
    \bottomrule
    \end{tabular}
    \end{adjustbox}
    \label{tab:ablation}
\end{table}
\vspace{-3mm}
We compare the efficiency of three saliency detectors: gradient-based, attention-based, and random baseline.
For the gradient-based detector, we follow the method used in Co-Mixup~\cite{kim2021co}, and random baseline refers to a random selection of tokens.
In Table~\ref{tab:saliency_map}, we report the average latency of saliency detection per iteration in our CIFAR-100 experiment setting with a batch size of 128.
Compared to the gradient-based method, attention-based saliency detection with 1-step rollout achieves approximately $\times15$ speed-up while demonstrating even better performance.

\subsection{Component Analysis}
% In this section, we analyze the effect of each component in Horizontal TokenMixup and Vertical TokenMixup so as to answer \textbf{Q2}.
% To demonstrate the robustness of our method, we conduct a series of sensitivity tests with respect to  key hyperparameters, $\tau$, $\rho$, and $\kappa$.

% $\tau$ refers to the sample difficulty threshold, where setting low $\tau$ leads to overestimation of sample difficulty, and \textit{vice versa}.
% From left Table~\ref{tab:tau} we found $\tau = 2.0$ to be optimal, and by setting $\tau = 0$, all samples were regarded difficult and none of the instanced had TokenMixup applied. 
% In the other extreme with infinite $\tau$, \textit{i.e.} TokenMixup applied to all instances, performance of \answerTODO{} was recorded.
% This represents the case where the ScoreNet takes no effect.

% $\rho$, on the other hand, refers to the saliency difference threshold.
% That is, $\rho$ controls the minimum amount of saliency gain required for a token to be replaced.
% By setting $\rho = 0$, tokens are mixed in a way that maximizes total saliency.
% If $\rho$ is maximal, no tokens are mixed, as shown in middle Table~\ref{tab:rho}.

% Finally, $\kappa$ is the number of tokens to be pooled from each previous layers when VTM is adopted.
% We did not observe specific trends or patterns by controlling $\kappa$.
% However, we could see that model performance is robust to this hyperparameter.
In this section, we analyze the effect of each component in Horizontal TokenMixup and Vertical TokenMixup so as to answer \textbf{Q2}.
In Table~\ref{tab:ablation}, we provide ablation studies on components of HTM: Sample difficulty assessment (ScoreNet), Optimal assignment, Token-level mixup, and Saliency-based label reassignment.
By ablating ScoreNet, all instances are mixed regardless of sample difficulty; note, ScoreNet still receives supervision signals.
By getting rid of optimal assignment, mixup pairs are not assigned via Hungarian matching; pairs are randomly selected.
Token-level mixup and label reassignment is relevant to the method by which tokens are selected.
Then, by ablating mixup, we randomly select tokens to replace, while label reassignment ablation derives new labels based on the number of tokens that have been replaced.
% In Table~\ref{tab:random_baseline}, we further compare the performance of HTM and VTM with and without saliency-based mixup.
% It is observed that mixing tokens with respect to saliency difference is better than randomly selecting tokens to replace.
% From the table, we can observe that optimal pair assignment and saliency-based label reassignment contributes the most in terms of accuracy.
We also conducted sensitivity tests on key hyperparameters $\tau$, $\rho$, and $\kappa$, but tables and discussions were moved to the supplement due to spatial constraints.

\subsection{What TokenMixup Learns}
\label{sec:analysis_learn}
\begin{table}

\centering
\caption{\footnotesize \textbf{Mixup combinations.}  CCT performance on CIFAR-100 is compared for mixup method combinations. $^*$ denotes values reported in the original paper. Best performances among comparable settings are highlighted.}
\label{tab:manifold}
\begin{adjustbox}{width=0.7\textwidth}
\begin{tabular}{l | c}
\toprule
\textbf{Mixup Combination} & \textbf{CIFAR-100 Accuracy} \\
\midrule
CCT-7/3x1 & 76.80 \\
CCT-7/3x1 + Manifold Mixup & 76.72 \\
CCT-7/3x1 + \textbf{Horizontal TM} & 77.34 \\
CCT-7/3x1 + \textbf{Vertical TM} & \cellcolor{yellow!25} 78.18  \\
CCT-7/3x1 + Input Mixup \& Cutmix $^*$ & 82.87 \\
CCT-7/3x1 + Input Mixup \& Cutmix + Manifold Mixup & 80.08 \\
CCT-7/3x1 + Input Mixup \& Cutmix + \textbf{Horizontal TM} & \textbf{83.56} \\
CCT-7/3x1 + Input Mixup \& Cutmix + \textbf{Vertical TM} & \textbf{83.54} \\
CCT-7/3x1 + Input Mixup \& Cutmix + \textbf{HTM} + \textbf{VTM} & \cellcolor{yellow!25}\textbf{83.57} \\

\bottomrule
\end{tabular}
\end{adjustbox}

\end{table}

\vspace{-3mm}
Here, we answer our research question \textbf{Q3} by analyzing what our mixup methods are learning.
% In the previous sections, we have discussed a curriculum learning perspective of Horizontal TokenMixup.
We demonstrate a curriculum learning perspective in terms of HTM, and compare our method with a direct mixup alternative, manifold mixup~\cite{verma2019manifold}.
% its property qualitatively in the following subsection.
% Moreover, we compare HTM with a direct mixup alternative, Manifold Mixup~\cite{verma2019manifold} by applying it to our baseline transformer models.

\vspace{-3mm}
\begin{wrapfigure}{r}{70mm}
\vspace{-0.9cm}
\begin{center}
\includegraphics[width=70mm]{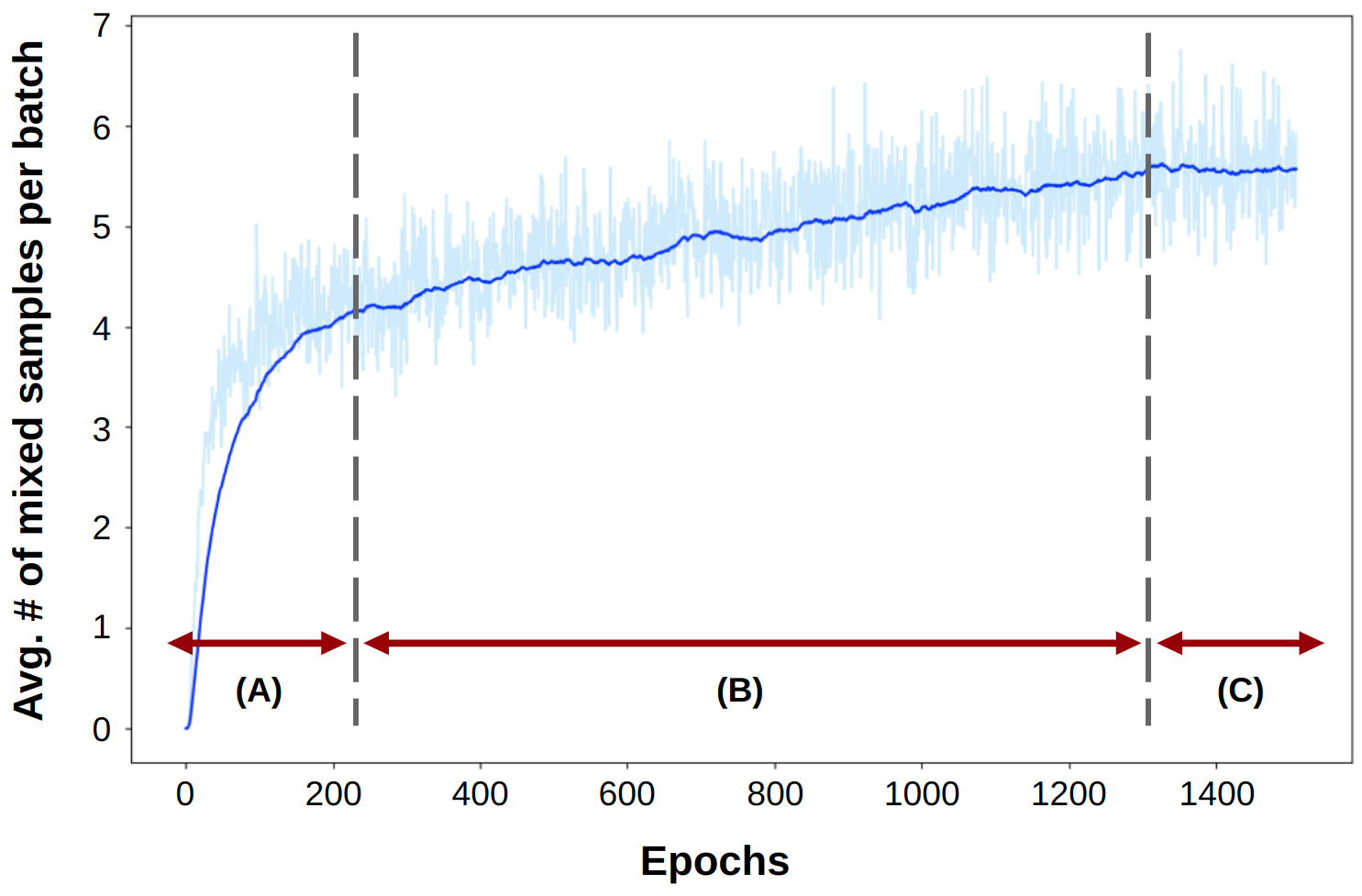}
\end{center}
\vspace{-5mm}
\caption{\footnotesize \textbf{Trend in the number of mixed instances.}}
\label{fig:curriculum}
\vspace{-0.8cm}
\end{wrapfigure} 
\paragraph{A Curriculum Learning perspective.}
In Figure~\ref{fig:curriculum} we plot the trend in the number of samples per batch TokenMixup is activated based on ScoreNet.
In the first few epochs (regime (A)), the number of mixed samples is relatively low, as there is less need for augmentation in the early phase of learning.
But as the learnable parameters warm up, the number increases abruptly.
In regime (B), the average number of mixed tokens slowly increases as model performance steadily improves.
In the last learning phase (regime (C)), mixup frequency converges.

\vspace{-2mm}
\paragraph{Comparison with Manifold Mixup.}
TokenMixup is applied to the intermediate tokens, similarly to manifold mixup~\cite{verma2019manifold} which is an input mixup~\cite{zhang2017mixup} method applied to intermediate features.
By using manifold mixup, transformer tokens are mixed via linear interpolation, instead of the hard replacement of tokens as in TokenMixup.
In Table~\ref{tab:manifold}, our methods are compared with manifold mixup, along with various combinations of original mixup methods (\textit{i.e.}, input mixup and cutmix).
The best setting in the CCT paper adopted input mixup and cutmix. 
By applying HTM and VTM on top of that setting, we achieved state-of-the-art performance.
On the other hand, by applying manifold mixup, we observed severe deterioration in accuracy.
We conjecture that soft mixing of transformer tokens leads to over-smoothing of features, which may be inadequate for self-attention layers.
See the supplement for visual aid.

\vspace{-2mm}
\section{Conclusion}
\vspace{-3mm}
We proposed TokenMixup for transformers, which adopts the attention map in place of computationally heavy gradient-based saliency detectors.
By adopting the attention as the saliency map, we achieve $\times15$ speed-up compared to the gradient-based method with even better performance.
We also introduce a novel perspective of curriculum learning to mixup, which enables adaptive augmentation based on sample difficulty and training schedule.
To add diversity, Vertical TokenMixup is introduced, which mixes tokens from different layers for multi-scale feature augmentation within a single sample.
With TokenMixup, we achieve state-of-the-art performance on CIFAR-100, and improve baselines by significant margins on CIFAR-10 and ImageNet-1K.

\begin{ack}
This work was partly supported by ICT Creative Consilience program (IITP-2022-2020-0-01819) supervised by the IITP; the National Supercomputing Center with
supercomputing resources including technical support (KSC-2022-CRE-0100) and Kakao Brain Corporation.
\end{ack}

% 명품인재  / 카카오브레인 / KISTI  KSC-2021-CRE-0299

\bibliographystyle{unsrt}
{\small
\bibliography{reference}
}
\newpage

%%%%%%%%%%%%%%%%%%%%%%%%%%%%%%%%%%%%%%%%%%%%%%%%%%%%%%%%%%%%
\section*{Checklist}

%%% BEGIN INSTRUCTIONS %%%
% The checklist follows the references.  Please
% read the checklist guidelines carefully for information on how to answer these
% questions.  For each question, change the default \answerTODO{} to \answerYes{},
% \answerNo{}, or \answerNA{}.  You are strongly encouraged to include a {\bf
% justification to your answer}, either by referencing the appropriate section of
% your paper or providing a brief inline description.  For example:
% \begin{itemize}
%   \item Did you include the license to the code and datasets? \answerYes{See Section~\ref{gen_inst}.}
%   \item Did you include the license to the code and datasets? \answerNo{The code and the data are proprietary.}
%   \item Did you include the license to the code and datasets? \answerNA{}
% \end{itemize}
% Please do not modify the questions and only use the provided macros for your
% answers.  Note that the Checklist section does not count towards the page
% limit.  In your paper, please delete this instructions block and only keep the
% Checklist section heading above along with the questions/answers below.
%%% END INSTRUCTIONS %%%

%%%%%%%%%%%%%%%%%%%%%%%%%%%%%%%%%%%%%%%%%%%%%%%%%%%%%%%%%%%%
\begin{enumerate}

\item For all authors...
\begin{enumerate}
  \item Do the main claims made in the abstract and introduction accurately reflect the paper's contributions and scope?
    \answerYes{See Section~\ref{section:method}.}
  \item Did you describe the limitations of your work?
    \answerYes{See the supplement}
  \item Did you discuss any potential negative societal impacts of your work?
    \answerYes{See the supplement}
  \item Have you read the ethics review guidelines and ensured that your paper conforms to them?
    \answerYes{}
\end{enumerate}

\item If you are including theoretical results...
\begin{enumerate}
  \item Did you state the full set of assumptions of all theoretical results?
    \answerNA{}
 \item Did you include complete proofs of all theoretical results?
    \answerNA{}
\end{enumerate}

\item If you ran experiments...
\begin{enumerate}
  \item Did you include the code, data, and instructions needed to reproduce the main experimental results (either in the supplemental material or as a URL)? \answerYes{}
  \item Did you specify all the training details (e.g., data splits, hyperparameters, how they were chosen)?  \answerYes{See Section~\ref{section:experiments} and the supplement for technical details}
  \item Did you report error bars (e.g., with respect to the random seed after running experiments multiple times)? \answerNA{Our experiment settings do not conduct each experiment multiple times.}
  \item Did you include the total amount of compute and the type of resources used (e.g., type of GPUs, internal cluster, or cloud provider)?  \answerYes{See Section~\ref{section:experiments} and the supplement for technical details.}
\end{enumerate}

\item If you are using existing assets (e.g., code, data, models) or curating/releasing new assets...
\begin{enumerate}
  \item If your work uses existing assets, did you cite the creators?
    \answerYes{See the supplement or Section~\ref{section:experiments} baseline.}
  \item Did you mention the license of the assets?
    \answerYes{See the supplement}
  \item Did you include any new assets either in the supplemental material or as a URL?
    \answerYes{See the supplement}
  \item Did you discuss whether and how consent was obtained from people whose data you're using/curating?
    \answerYes{See the supplement}
  \item Did you discuss whether the data you are using/curating contains personally identifiable information or offensive content?
    \answerYes{See the supplement}
\end{enumerate}

\item If you used crowdsourcing or conducted research with human subjects...
\begin{enumerate}
  \item Did you include the full text of instructions given to participants and screenshots, if applicable?
    \answerNA{}
  \item Did you describe any potential participant risks, with links to Institutional Review Board (IRB) approvals, if applicable?
    \answerNA{}
  \item Did you include the estimated hourly wage paid to participants and the total amount spent on participant compensation?
    \answerNA{}
\end{enumerate}

\end{enumerate}

\end{document}